\documentclass[11pt]{article}
\usepackage{coling2018}
\usepackage{times}
\usepackage{lmodern}
\usepackage{url}
\usepackage{bold-extra}
\usepackage{latexsym}
\usepackage{graphicx}
\usepackage{amsmath,amssymb,amsthm,dsfont,color,enumitem}
\usepackage{url}
\usepackage{subcaption}
\usepackage{bold-extra}
\usepackage{bm}
\usepackage{xcolor}
\usepackage{comment}
\usepackage{multirow}
\usepackage[title]{appendix}
\usepackage[ruled,vlined,noend,linesnumbered]{algorithm2e}

\newcommand{\eq}{\phantom{{}+{}}=\phantom{{}+{}}}
\newcommand{\proptoo}{\phantom{{}+{}}\propto\phantom{{}+{}}}

\theoremstyle{plain}
\newtheorem{theorem}{Theorem}
\newtheorem{definition}{Definition}

\newcommand{\mltm}[0]{multilingual topic model}
\newcommand{\mltms}[0]{multilingual topic models}

\newcommand{\es}[0]{\textsc{es}}
\newcommand{\ar}[0]{\textsc{ar}}
\newcommand{\fa}[0]{\textsc{fa}}
\newcommand{\zh}[0]{\textsc{zh}}
\newcommand{\ru}[0]{\textsc{ru}}
\newcommand{\en}[0]{\textsc{en}}
\newcommand{\deltadtds}[0]{\left(\delta_{d_{\ell 2}}\right)_{d_{\ell 1}}}
\newcommand{\rdeltadtds}[0]{\left(\widetilde{\delta}_{d_{\ell 2}}\right)_{d_{\ell 1}}}
\newcommand{\gf}[1]{\Gamma\left(#1\right)}

\newcommand{\cnpmi}[0]{\textsc{cnpmi}}

\newcommand{\ds}{{d_{\ell 1}}}
\newcommand{\dt}{{d_{\ell 2}}}

\newcommand{\vt}[1]{\mathbf{#1}}

\newcommand{\hardlink}[0]{\textsc{hardlink}}
\newcommand{\voclink}[0]{\textsc{voclink}}
\newcommand{\softlink}[0]{\textsc{softlink}}
\newcommand{\ted}[0]{\textsc{ted}}
\newcommand{\gv}[0]{\textsc{gv}}

\newcommand{\wikiinco}[0]{\textsc{wiki-inco}}
\newcommand{\wikipaco}[0]{\textsc{wiki-paco}}
\newcommand{\lda}[0]{\textsc{lda}}
\newcommand{\lis}[0]{\textsc{lis}}

\allowdisplaybreaks

\title{Learning Multilingual Topics from Incomparable Corpora}

\author{Shudong Hao \\
  Computer Science \\
  University of Colorado \\
  Boulder, CO, USA \\
  {\url{shudong@colorado.edu}} \\\And
  Michael J. Paul \\
  Information Science \\
  University of Colorado \\
  Boulder, CO, USA \\
  {\url{mpaul@colorado.edu}} \\}

\date{}

\begin{document}
\maketitle
\begin{abstract}
Multilingual topic models enable crosslingual tasks
by extracting consistent topics from multilingual corpora.
Most models require parallel or comparable training corpora,
which limits their ability to generalize.
In this paper,
we first demystify the knowledge transfer mechanism behind multilingual topic models
by defining an alternative but equivalent formulation.
Based on this analysis,
we then relax the assumption of training data required by most existing models,
creating a model that only requires a dictionary for training.
Experiments show that our new method effectively learns
coherent multilingual topics from partially and fully \textit{incomparable}
corpora with limited amounts of dictionary resources.
\end{abstract}

\section{Introduction}
\label{sec:intro}

Multilingual topic models provide an overview of document structures in multilingual corpora, by learning language-specific versions of each topic (Figure~\ref{fig:multitopics}). 
Their simplicity, efficiency and interpretability make models from this family 
popular for various crosslingual tasks, \textit{e.g.,} feature extraction~\cite{LiuDM15}, cultural difference discovery~\cite{ShutovaSGLN17,GutierrezSLMG16}, 
translation detection~\cite{KrstovskiSK16,KrstovskiS16},
and others~\cite{BarrettKS16,AgicJPASS16,HintzB16}.

Typical probabilistic multilingual topic models
are based on
Latent Dirichlet Allocation (\textsc{lda},~\newcite{BleiNJ03}),
adding supervision on connections between languages.
Most models achieve this by making strong assumptions
on the training data---they
either require a \textit{parallel corpus} that has sentence-aligned documents in different languages (\textit{e.g.,} EuroParl, \newcite{europarl}),
or a \textit{comparable corpus} that has documents of similar content
(\textit{e.g.,} Wikipedia articles paired across languages).
These training requirements limit the usage
of such models:
an adequately large parallel corpus is difficult to obtain, particularly for low-resource languages.
For example, only $300$ languages are available on Wikipedia,\footnote{\url{https://meta.wikimedia.org/wiki/List_of_Wikipedias}} 
and only $250$ languages have more than $1{,}000$ articles.
Another common choice for parallel corpus in multilingual research, the Bible,
is available in $2{,}530$ languages~\cite{AgicHS15}.\footnote{Reported by United Bible Societies at \url{https://www.unitedbiblesocieties.org/}}
However, studies show that its archaic themes and small corpus size ($1{,}189$ chapters)
can limit performance~\cite{HaoBGP18,MoritzB17}.
Therefore,
the requirement of parallel/comparable corpora for multilingual topic models
limits their usage in many situations.

Another line of research focuses on 
using multilingual dictionaries as supervision~\cite{MaN17,GutierrezSLMG16,LiuDM15,JagarlamudiD10,BoydGraberB09}.
In contrast to parallel corpora,
dictionaries are widely available and often easy to obtain.
\textsc{PanLex}, a free online dictionary database, for example,
covers $5{,}700$ languages and more than one billion dictionary entries~\cite{KamholzPC14,BaldwinPC10}.\footnote{\url{https://panlex.org/}}
Thus, a multilingual topic model built on a dictionary rather than a parallel corpus
is potentially applicable to more languages.

A dictionary also allows for training on \textit{incomparable} corpora---documents in different languages that are from different sources without direct connections---which have had less research on learning consistent topics.
With a dictionary, a natural question is how to efficiently utilize the semantic information it carries  so that a topic model can
produce multilingually coherent topics. 
This work considers an alternative formulation of a dictionary-based topic model, one that borrows the structure of models used with comparable corpora, but uses a dictionary-based metric to learn connections between documents, instead of explicit connections from a comparable corpus.
The main contributions of this work are:
\vspace{-0.6em}
\begin{itemize}[itemsep=-1ex]
	\item We summarize existing related work in Section~\ref{sec:prelim} and propose a new formulation of multilingual topic models based on crosslingual transfer learning in Section~\ref{sec:proposed}. This new formulation explicitly shows the knowledge transfer mechanism during the generative process.
	\item Based on this new formulation, in Section~\ref{sec:softlink} we generalize existing multilingual topic models and relax the assumptions of parallel/comparable datasets. Our approach requires only a dictionary, and is empirically shown to perform well even with only limited amounts of available entries.
	\item We evaluate our new model on five languages from different language families in Section~\ref{sec:expr}. Our proposed model learns multilingually coherent topics and yields around a $25\%$ relative improvement in crosslingual classification performance.
\end{itemize}

\section{Multilingual Topic Models}
\label{sec:prelim}

\begin{figure}
	\centering
	\includegraphics[width=0.85\linewidth]{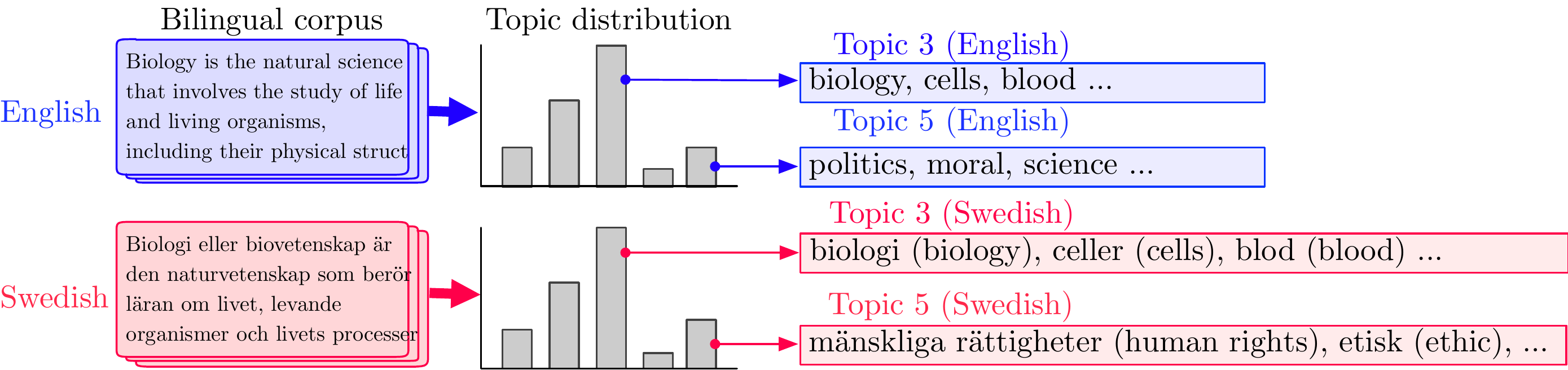}
	\caption{Multilingual topic models produce topics where each language has its own version.}
	\label{fig:multitopics}
\end{figure}

Multilingual topic models generate $K$ topics from a corpus consisting of multiple languages;
each topic has a {version} specific to each language in the corpus (Figure~\ref{fig:multitopics}).
From a human's view,
a coherent multilingual topic should talk about the same thing regardless of the language;
from a machine's view,
the success of \mltms{} depends on the inferred topics being consistent across languages.
For example,
given an English-Swedish bilingual topic $\phi_k^{(\en{},\textsc{sv})}$,
the probability of an English word \textit{island} and that of its translation in Swedish, \textit{\"{o}}, should be similar, \textit{i.e.,}
$\Pr\left(island_{\en{}}|\phi_k^{(\en{},\textsc{sv})}\right) \approx \Pr\left(\ddot{o}_{\textsc{sv}}|\phi_k^{(\en{},\textsc{sv})}\right)$.
Most multilingual topic models extend \lda{} with
one or both of two types of ``link'' information:
document translations and word translations.

\paragraph{Document Links.}
The polylingual topic model \cite{MimnoWNSM09,NiSHC09} assumes that during the generative process, a topic distribution $\theta_{\mathbf{d}}$ generates a tuple of comparable documents  in different languages, \textit{i.e.,} $\mathbf{d}=\left(d^{(\ell_1)},\ldots,d^{(\ell_L)}\right)$
and each language $\ell$ has its own topic-word distributions, $\phi^{(\ell)}_k$.
This model has been widely used \cite{VulicSM13,PlattTY10,SmetM09},
but it requires a parallel/comparable corpus in order to link documents.

\paragraph{Vocabulary Links.}
Another type of model uses word translations~\cite{JagarlamudiD10,BoydGraberB09} rather than linking documents. 
A multilingual dictionary is used to
construct a tree structure where each internal node contains word translations, and applies hyper-Dirichlet type I distributions to generate words~\cite{AndrzejewskiZC09,minka,dennis1991}.
For each topic $k$,
a distribution from root $r$ to all the internal nodes $i$ is drawn by $\phi_{k,r}\sim\mathrm{Dir}(\beta_r)$,
and then a distribution from $i$ to a leaf node is drawn by $\phi^{(\ell)}_{k,i}\sim\mathrm{Dir}\left(\beta_i^{(\ell)}\right)$.
A word $w^{(\ell)}$ in language $\ell$ is drawn from a product of the two multinomial distributions by $w^{(\ell)}\sim\mathrm{Mult}\left(\phi_{k,r}\cdot\phi^{(\ell)}_{k,i}\right)$.

\paragraph{Variations.}
Many variations of these ideas have been proposed to deal with non-parallel corpus.
\newcite{HeymanVM16} proposed \textsc{C-BiLDA},
which distinguishes between shared and non-shared topics
across languages, based on a document links model.
The model, however,
 requires a comparable dataset that
provides document links between languages.
A variation proposed by~\newcite{MaN17} deals with non-parallel corpora.
This model is essentially a modified version of \newcite{JagarlamudiD10} and \newcite{BoydGraberB09},
so we consider this work to be another vocabulary links model.
Other models have been proposed for very specific situations that needs additional supervision.
For example, \newcite{KrstovskiSK16} requires scientific article section alignments, and \newcite{GutierrezSLMG16} requires Part-of-Speech (\textsc{pos}) taggers, which are not always available for all languages. Without \textsc{pos} taggers, this model is equivalent to vocabulary links.
In our work, we focus on the standard document links and vocabulary links models, which are the most generalizable models.

\section{Document Links: A Crosslingual Transfer Perspective}
\label{sec:proposed}

Before we introduce our new approach,
we first present an alternative understanding of the document links model
from the perspective of crosslingual transfer learning.
In multilingual topic models,
``knowledge'' refers to
word distributions for a topic in a language $\ell$,
and we study how multilingual topic models
transfer this knowledge from one language to another
so that the model provides semantically coherent topics
that are consistent across languages.

In the standard document links model,
a ``link'' between a document $d_{\ell_1}$ in language $\ell_1$
and $d_{\ell_2}$ in $\ell_2$ indicates that
they are translations or closely comparable.
In this model, the topic assignments for both documents are independently generated
from the same distribution, $\theta_{d_{\ell 1},d_{\ell 2}}$.
Thus, the joint likelihood of document links model is:
\begin{align}
\Pr\left(\mathbf{w}_{d_{\ell 1}},\mathbf{z}_{d_{\ell 1}},\mathbf{w}_{d_{\ell 2}},\mathbf{z}_{d_{\ell 2}} | \alpha,\beta\right),\label{joint}
\end{align}
where $\mathbf{w}_{d_{\ell}}$ and $\mathbf{z}_{d_{\ell}}$ are the word tokens and topic assignments of document $d_\ell$.
We refer this formulation as the \textbf{joint generative model},
since the topics and words of $d_{\ell_1}$ and $d_{\ell_2}$
are generated \textit{simultaneously}.

The simultaneousness of this model formulation, in which
both languages generate topics jointly,
masks the knowledge transfer process.
To highlight this process, and to help us generalize the model in the next section,
we define an alternative formulation
in which $d_{\ell_1}$ and $d_{\ell_2}$
are generated \textit{sequentially}.

Assume the topics of $d_{\ell_1}$ have already been generated from $\theta_{d_{\ell 1}}\sim\mathrm{Dir}(\alpha)$,
and $\mathbf{n}_{d_{\ell 1}}\in\mathbb{N}^{K}$ is a vector
of topic counts in $d_{\ell_1}$.
In our alternative formulation, the generation of topics of $d_{\ell_2}$
depends on $d_{\ell_1}$ by
$\theta_{d_{\ell 2}}\sim\mathrm{Dir}(\alpha + \mathbf{n}_{d_{\ell 1}})$,
where the prior $\alpha + \mathbf{n}_{d_{\ell 1}}$ encourages the 
distribution $\theta_{d_{\ell 2}}$
to be similar to $\theta_{d_{\ell 1}}$.
This formulation can go the other way,
\textit{i.e.,} generating $d_{\ell_2}$ first,
and then $d_{\ell_1}$.
The combined likelihood of this formulation is:
\begin{align}
\label{sep}
\Pr\left(\mathbf{w}_{d_{\ell 1}},\mathbf{z}_{d_{\ell 1}} | \mathbf{w}_{d_{\ell 2}},\mathbf{z}_{d_{\ell 2}}, \alpha,\beta\right)
\cdot
\Pr\left(\mathbf{w}_{d_{\ell 2}},\mathbf{z}_{d_{\ell 2}} | \mathbf{w}_{d_{\ell 1}},\mathbf{z}_{d_{\ell 1}}, \alpha,\beta\right),
\end{align}
and we refer to this formulation as the \textbf{conditional generative model}.

This alternative formulation explicitly
shows the knowledge transfer process across languages
by shaping the topic parameters for $\ell_2$
to be similar to that of the other language $\ell_1$, and vice versa.
In this formulation,
the likelihood of the conditional generative model
is different from the joint generative model.
In fact, this is an instance of
{\em pseudolikelihood}~\cite{besag1975statistical,LeppaahoPRC17},
where the joint likelihood of the two documents
is approximated as the product of each document's
conditional likelihood given the other, \textit{i.e.,}
$\Pr(d_{\ell_1}, d_{\ell_2}) \approx \Pr(d_{\ell_1}|d_{\ell_2})\cdot\Pr(d_{\ell_2}|d_{\ell_1})$.
As \newcite{LeppaahoPRC17} suggests,
pseudolikelihood is not a numerically accurate approximation to the joint likelihood;
Theorem~\ref{th:pll} below, however,
states that this formulation yields exactly the same posterior estimations of $\theta$ and $\phi$.

\begin{theorem}\label{th:pll}
	The conditional generative model with document links 
	yields the same posterior estimator to the joint generative model
	using collapsed Gibbs sampling.
\end{theorem}

\begin{proof}
See Appendix.
\end{proof}

\section{Generalizing Document Links}
\label{sec:softlink}

Obtaining parallel corpora for training the document links model
is very demanding, particularly for low-resource languages.
Therefore,
as the second major contribution in this paper,
we generalize the document links model
using the formulation described above
to require only a bilingual dictionary.

\subsection{From Hard Links to Soft Links}

Following the above discussion,
we introduce our method assuming the directionality from language $\ell_1$ to $\ell_2$.
We generalize the model above by rewriting the generation of the distribution $\theta_{d_{\ell 2}}$:
\begin{equation}
\theta_{d_{\ell 2}} \phantom{=}\sim\phantom{=} \mathrm{Dirichlet}\left(\alpha+\delta_{d_{\ell 2}}\cdot\mathbf{N}^{(\ell_1)}\right),
\end{equation}
where $\mathbf{N}^{(\ell_1)}\in\mathbb{N}^{|D^{(\ell_1)}|\times K}$ is the matrix of topic counts per document with $K$ topics in the corpus $D^{(\ell_1)}$ of language $\ell_1$.
This is equivalent to the above document links model when $\delta_{d_{\ell 2}}\in\mathbb{R}^{|D^{(\ell_1)}|}$ is an indicator vector that has value $1$ for the corresponding parallel document $d_{\ell_1}\in D^{(\ell_1)}$ and $0$ elsewhere.
We refer this as {\bf hard links} (\hardlink{}), where each document $d_{\ell_2}\in D^{(\ell_2)}$ is informed by exactly one document $d_{\ell_1}$, and this link is known \textit{a~priori} from a parallel corpus. 

\begin{figure}
	\centering
	\includegraphics[width=\linewidth]{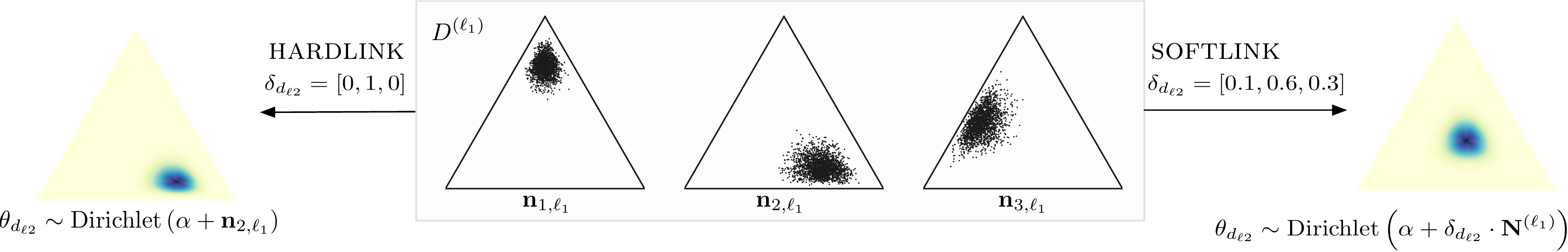}
	\caption{An illustration of how topic knowledge is transferred across languages through \hardlink{} and \softlink{}. To generate observations in $d_{\ell_2}$, both models uses topics in $\ell_1$ as prior knowledge to shape the Dirichlet prior for $d_{\ell_2}$. This transfer happens in \hardlink{} by aligned documents in a comparable corpus, while \softlink{} uses a generalized transfer distribution $\delta$.}
	\label{fig:transfer-dir}
\end{figure}

We create {\bf soft links} (\softlink{}) by
relaxing the assumption that $\delta_{d_{\ell 2}}$
is an indicator vector,
instead allowing $\delta_{d_{\ell 2}}$ to be
any {\em distribution} over documents in $D^{(\ell_1)}$,
a mixture of potentially multiple documents in language $\ell_1$
to inform parameters for a document $d_{\ell_2}$ in language $\ell_2$. 
We refer this distribution as the \textbf{transfer distribution}.
The Dirichlet prior for document ${d_{\ell_2}}$
contains topic knowledge $\mathbf{N}^{(\ell_1)}$ transferred from corpus $D^{(\ell_1)}$,
encouraging $\theta_{d_{\ell 2}}$ to be proportionally similar to documents
in $D^{(\ell_1)}$.
Figure~\ref{fig:transfer-dir} illustrates this process.

\subsection{Defining the Transfer Distribution}
\label{sec:threshold}

The transfer distribution of document $d_{\ell_2}$
indicates how much knowledge should be transferred from every document $d_{\ell_1}\in D^{(\ell_1)}$.
Intuitively, if $d_{\ell_1}$ and $d_{\ell_2}$ have a large amount of overlapping word translations,
their topics should be similar as well.
Therefore,
we define the values of $\delta$ based on the similarity of document pairs using a bilingual dictionary.
Specifically, for a document $d_{\ell_2}\in D^{({\ell_2})}$ ,
the transfer distribution of $d_{\ell_2}$, denoted as $\delta_{d_{\ell 2}}$, is a normalized vector of size $|D^{(\ell_1)}|$,
\textit{i.e.,} the size of corpus $D^{(\ell_1)}$.
Each cell in $\delta_{d_{\ell 2}}$ corresponds to a document $d_{\ell_1}\in D^{(\ell_1)}$, defined as: 
\begin{align}
\deltadtds\proptoo \frac{|\left\{w_{\ell_1}\right\}\cap\left\{w_{\ell_2}\right\}|}{|\left\{w_{\ell_1}\right\}\cup\left\{w_{\ell_2}\right\}|}, \phantom{=+=}
\forall \ w_{\ell_1} \in d_{\ell_1} w_{\ell_2} \in d_{\ell_2},
\end{align}
where $\{w_\ell\}$ contains all the word types that appear in document $d_{\ell}$,
and $\left\{w_{\ell_1}\right\}\cap\left\{w_{\ell_2}\right\}$ indicates all word pairs $(w_{\ell_1},w_{\ell_2})$ that can be found in a dictionary as translations.
In other words, $\deltadtds$ is the proportion of words in the document pair $\left(d_{\ell_1},d_{\ell_2}\right)$ that are translations of each other.

In practice, a dense transfer distribution is computationally inefficient and is less meaningful than a sparse distribution, as it becomes approximately uniform due to the large size of the corpus.
The transfer distribution should be more heavily concentrated on documents
with higher word-level translation probabilities,
while reducing the noise negatively transferred from those with low probabilities.
To this end,
we propose two approaches to help transfer distributions more efficiently
focus on specific documents.

\subsubsection{Static Focusing: a Threshold Method}

The first method is to focus the distribution on the highest values 
such that values below a threshold are set to $0$, while the remaining values are renormalized to sum to $1$. 
The modified distribution is thus:
\begin{align}
	\rdeltadtds \proptoo \mathds{1}\bigg\{\deltadtds > \pi \cdot \max\left(\delta\right)\bigg\}\cdot\deltadtds
\end{align}
where $\mathds{1}$ is an indicator function,
and $\pi\in[0,1]$ is the \textbf{focal threshold},
a fixed parameter that adjusts the threshold. 
The threshold is defined with respect to the maximum value of $\delta$.
A \textit{corpus-wise} threshold chooses $\max(\delta)$ from
all the $\delta_{\dt}$ in $D^{(\ell_2)}$ globally,
while we also consider a \textit{document-wise}
threshold for each document, $\pi \cdot \max\left(\delta_{\dt}\right)$.
We refer these two manners as the \textbf{selection scope}.

\begin{figure}[t]
	\centering
	\includegraphics[width=0.9\linewidth]{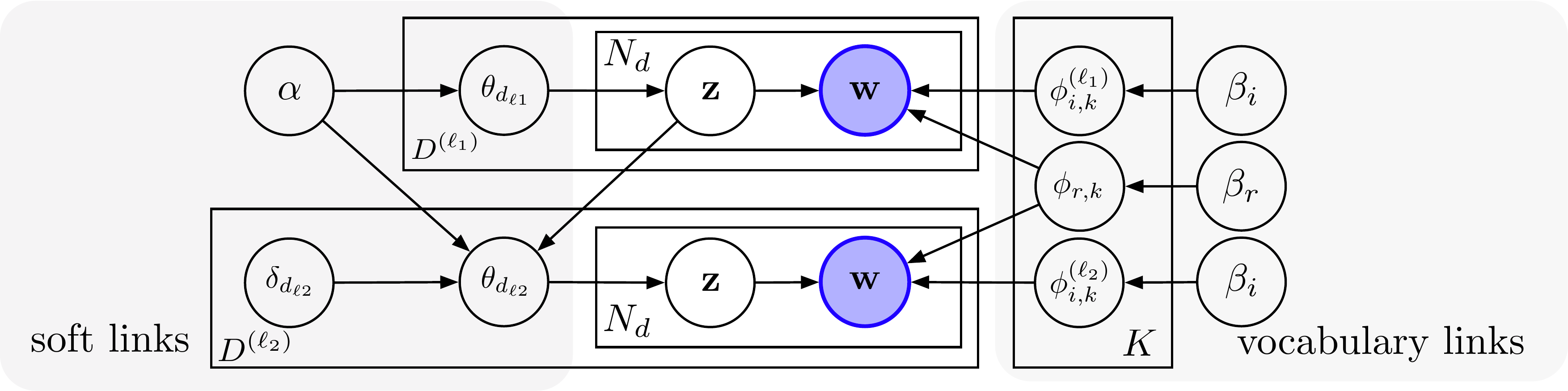}
	\caption{Plate notation of a \mltm{} using soft links and vocabulary links.}
	\label{fig:plate}
\end{figure}

\subsubsection{Dynamic Focusing: an Annealing Method}
\label{sec:annealing}

Static focusing treats transfer distributions $\delta$ as fixed parameters
during sampling,
and it is difficult to decide how sparse
a transfer distribution should be to achieve optimal performance.
Therefore, we propose dynamic focusing,
where we avoid choosing a specific focal threshold and selection scope.
Specifically, we adjust the transfer distribution during inference {dynamically},
beginning with a dense transfer distribution
and iteratively sharpening the distribution
using deterministic annealing~\cite{UedaN94,SmithE06,PaulD15}.

Assume at iteration $t$,
the transfer distribution for a document $d_{\ell_2}$ is denoted as $\delta_{d_{\ell 2}}^{(t)}$.
Then at iteration $t'$, we anneal its transfer distribution by 
$
	\left(\delta_{d_{\ell 2}}^{(t')}\right)_{d_{\ell 1}} \propto  \left(\delta_{d_{\ell 2}}^{(t)}\right)_{d_{\ell 1}}^{1/\tau}
$
where $\tau$ is a fixed temperature, which we set to $0.9$ in our experiments.
We start with non-focused transfer distributions,
and apply annealing at scheduling intervals during Gibbs sampling.

Designing an effective annealing schedule is critical.
We propose two schedules below. 

\paragraph{Fixed Schedule.}
The simplest schedule is to apply annealing
for all transfer distributions every $I$ iterations.
In our experiments,
we set $I=10$ (\textit{i.e.,} $t' = t+10$) and stop annealing after $400$ iterations.
A potential problem with fixed schedule is that
it can ``over-anneal'' the transfer distributions,
\textit{i.e.,} all the mass converges to only one document.

\paragraph{Adaptive Schedule.}
A robust multilingual topic model
should produce similar distributions over topics
for a pair of word translations $c=\left(w_{\ell_1}, w_{\ell_2}\right)$,
where we call $c$ a concept.
In other words, given a topic $k$,
the probability of expressing a concept $i$ in language $\ell_1$
should be similar to language $\ell_2$.
Thus, during iteration $t$,
we calculate $\varphi^{(\ell,t)}_c$,
the distribution over $K$ topics for each concept $c$ for each language $\ell$.
Using $\varphi^{(\ell,t)}_c$ as features and its language $\ell$ as labels,
we perform five-fold cross-validation by logistic regression
for all concepts $c$.
We define the average classification accuracy over the five folds
as the \textbf{language identification score} (\lis{}).
The lower the \lis{},
the better the model,
since a high \lis{} means the inferred distributions are inconsistent enough to
discriminate between languages.
This idea is related to adversarial training
between languages~\cite{ChenASWC16}.

During Gibbs sampling,
we calculate \lis{} after each iteration,
and average \lis{} every $I$ iterations.
We anneal all transfer distributions at iteration $t$
only if $\overline{\lis{}}_{t-I:t} > \overline{\lis{}}_{t-2I:t-I}$.
That is,
if the average \lis{} score during iteration $t-I$ and $t$
has been increasing since iteration $t-2I$ to $t-I$,
we treat this as a warning sign of increased \lis{} and thus anneal the transfer distributions.
As we sharpen $\delta$ by annealing,
knowledge transfer between languages becomes more specific.

\subsection{Modularity of Models}
Multilingual topic models can include the different types of information we described in Section~\ref{sec:prelim}: document links, vocabulary links, or both~\cite{HuZEB14}, while a model with neither is equivalent to \textsc{lda}. Document links can be either hard or soft, or a mix of both, as the only distinction is whether the transfer distribution is an indicator vector.
A complete model with both soft document links and vocabulary links is
shown in Figure~\ref{fig:plate}.
In Section~\ref{sec:expr}, we experiment with a combination of \softlink{} and \voclink{}.

\section{Experiments}
\label{sec:expr}

\subsection{Data} 

We use five corpora in five languages from different language groups:
Arabic (\ar{}, Semitic), Spanish (\es{}, Romance), Farsi (\fa{}, Indo-Iranian), Russian (\ru{}, Slavic), and Chinese (\zh{}, Sinitic).
Each language is paired with English (\en{}, Germanic),
and we train multilingual topic models on these language pairs individually.
All the corpora listed below are available at \url{http://opus.nlpl.eu/}.
For preprocessing,
we use stemmers to lemmatize and segment Chinese documents,
and then remove stop words and the most frequent $100$ word types for each language.
Refer to the appendix for additional details.

\paragraph{Training corpora.}
As in many multilingual studies~\cite{Ruder17},
we use Wikipedia as our training corpus for multilingual topic models,
and create two corpora,
\wikiinco{} and \wikipaco{} for each language pair $(\en{}, \ell)$.
For \wikiinco{}, we randomly select $2,000$ documents in each language
without any connections, so that no documents are translations of each other (an \textit{incomparable} corpus).
We also create a \textit{partially comparable} corpus, \wikipaco{},
which contains around $30\%$ comparable document pairs for each language pair.

\paragraph{Test corpora.}

We create two test corpora for each language pair $(\en{},\ell)$
from TED Talks 2013 (\ted{}) and Global Voices (\gv{}),
which provide categories for each document
that can be used as classification labels.
The first one, \ted{}+\ted{},
contains documents from \ted{} in both languages,
while the second one, \ted{}+\gv{}
contains English documents from \ted{}
and non-English documents from \gv{}.
After training a topic model,
we use $\phi^{(\en{})}$ and $\phi^{(\ell)}$
to infer topics from both languages.
For \ted{}+\ted{},
we choose the five most frequent labels in \ted{} as the label set
(\textit{technology}, \textit{culture},
\textit{science}, \textit{global issues}, and \textit{design});
for \ted{}+\gv{},
we replace \textit{global issues} and \textit{design} with \textit{business} and \textit{politics},
since the label set from \gv{} does not include \textit{global issues} and \textit{design}.

\paragraph{Dictionary.}
We use Wiktionary to extract word translations for \voclink{} and to calculate transfer distribution values $\delta$ for \softlink{}.
The dictionary is available at \url{https://dumps.wikimedia.org/enwiktionary/}.

\subsection{Inference Settings}

For each compared model,
we set the number of topics $K=25$. We run the Gibbs samplers for $1,000$ training iterations
and $500$ iterations to infer topic distributions on test corpora.
We set Dirichlet priors $\alpha=0.1$ and $\beta=0.01$ for \hardlink{} and \softlink{}.
For \voclink{},
we set $\beta_r=0.01$ for priors from root to internal nodes,
and $\beta_i=100$ from internal nodes $i$ to leaves, following~\newcite{HuZEB14}.

\subsection{Evaluation Metrics}

We evaluate each model in two ways. 
Experimental results below are averaged across all language pairs.

\subsubsection{Intrinsic Evaluation: Multilingual Topic Coherence}
Typical topic model evaluations include intrinsic and extrinsic measurements.
Intrinsic evaluation focuses on topic quality or coherence of the trained topics.
The most widely-used metric for measuring monolingual topic coherence is normalized pointwise mutual information~\cite{LauNB14,NewmanLGB10}.
\newcite{HaoBGP18} proposed crosslingual normalized pointwise mutual information (\cnpmi{}) by extending this idea to multilingual settings,
which correlates well with bilingual speakers' judgments on topic quality.

Given a bilingual topic $k$ in languages $\ell_1$ and $\ell_2$,
and a parallel reference corpus $\mathcal{R}^{(\ell_1,\ell_2)}$,
the \cnpmi{} of topic $k$ is calculated as:
\begin{equation}
	\cnpmi{}(\ell_1,\ell_2,k) \eq \frac{1}{C^2}\sum_{i,j}^{C}\frac{1}{\log\Pr\left(w_{i}^{(\ell_1)},w_{j}^{(\ell_2)}\right)}\cdot\log \frac{\Pr\left(w_{i}^{(\ell_1)},w_{j}^{(\ell_2)}\right)}{\Pr\left(w_{i}^{(\ell_1)}\right)\Pr\left(w_{j}^{(\ell_2)}\right)}
\end{equation}

\noindent
where $C$ is the cardinality of a topic, \textit{i.e.,}
the $C$ most probable words in the topic-word distribution $\phi_k^{(\ell)}$.
The co-occurrence probability of two words, $\Pr\left(w_{i}^{(\ell_1)},w_{j}^{(\ell_2)}\right)$,
is defined as the proportion of document pairs where both words appear.
In the results below,
we set $C=20$, and average the \cnpmi{} scores over $K=25$ topics for each model output.

To calculate \cnpmi{} scores,
we use $10,000$ document pairs from a held-out portion of Wikipedia.
\cnpmi{} is an intrinsic evaluation,
so it is only available for the training sets, \wikipaco{} and \wikiinco{}.

\subsubsection{Extrinsic Evaluation: Crosslingual Classification}

A successful multilingual topic model
should provide informative features
for crosslingual tasks.
To show that our model is beneficial to downstream applications,
we use crosslingual document classification to evaluate topic model performance. A high classification accuracy when testing on a different language from training indicates topic consistency across languages~\cite{HermannB14,KlementievTB12,SmetTM11}.

As in other studies on multilingual topic models,
we first train topic models on a bilingual corpus ${D}^{(\ell_1,\ell_2)}$,
and then use topic-word distributions $\phi^{(\ell_1)}$ and $\phi^{(\ell_2)}$
to infer document-topic distributions on {unseen} documents ${D'}^{(\ell_1)}$ and ${D'}^{(\ell_2)}$.
Thus,
a classifier is trained on $\theta_{d_{\ell 1}}$ with corresponding labels
where $d_{\ell 1}\in{D'}^{(\ell_1)}$,
and tested on $\theta_{d_{\ell 2}}$ where $d_{\ell 2}\in{D'}^{(\ell_2)}$, and vice versa.
In our experiments,
we use \wikipaco{} and \wikiinco{} to train topic models first,
and then perform inference on either \ted{}+\ted{} (both English and non-English documents from \ted{})
or \ted{}+\gv{} (English documents from \ted{} and non-English from \gv{}).
For each language pair,
we train multi-label classifiers using support vector machines (SVM) with five-fold cross-validation on documents in one language and test on the other. The F-1 scores reported below are micro-averaged over all labels.

\subsection{Baseline Comparison}
\label{sec:baseline}

We first compare \softlink{} with other models: \hardlink{}, which is expected to do well on the partially comparable corpus (\wikipaco{}) but poorly on the incomparable corpus (\wikiinco{}), and \voclink{}.
We additionally combine \softlink{}+\voclink{}.

\begin{figure}
	\centering
	\includegraphics[width=\linewidth]{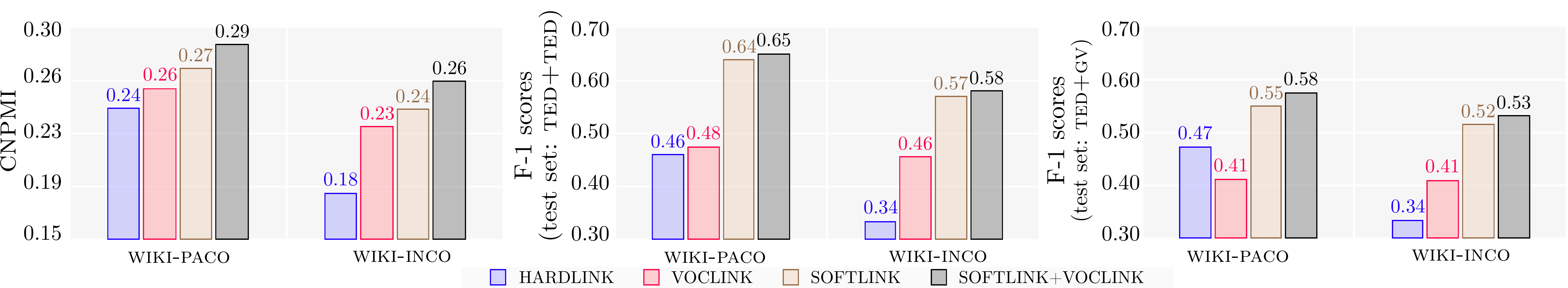}
	\caption{\softlink{} consistently outperforms other models on both topic quality evaluation (\cnpmi{}) and classification performance (F-1).}
	\label{fig:baselines}
\end{figure}

Figure~\ref{fig:baselines} shows the performance (both intrinsic and extrinsic) of all models. For the \softlink{} models, we used the optimal hyperparameter settings, but we compare other settings in Section~\ref{sec:compfoc}.

When the training corpus is partially comparable (\wikipaco{}),
all models can learn comparably coherent topics based on \cnpmi{} scores, though the \cnpmi{} of \hardlink{} is lower than all other models.
When the data is completely incomparable (\wikiinco{}),
\hardlink{} loses all connections between languages,
so as expected its topics are least coherent.
Similarly, when measuring classification performance, 
\hardlink{} is comparable to \voclink{} on \wikipaco{}, but much worse on \wikiinco{}, where it loses all information.
When the test set contains mostly parallel documents (\ted{}+\ted{}),
the F-1 scores are higher,
but when the test domain changes across languages (\ted{}+\gv{}),
the performance drops.

On the other hand,
\softlink{} consistently outperforms other models
regardless of training and test sets.
It seems that \softlink{} benefits from learning new connections between documents, even when part of the corpus contains direct links for training \hardlink{}.
It is also interesting that \softlink{} uses the same dictionary resource as \voclink{}, but has a relative performance increase around $25\%$.
It seems \softlink{} can more efficiently utilize lexical information
in a dictionary. 
We explore this relationship more in Section~\ref{sec:dictsize}.

Finally, we observe that combining \softlink{}+\voclink{} provides a performance boost over \softlink{} in all cases, though the increase is small.

\subsection{Comparison of Focusing Methods}
\label{sec:compfoc}

\begin{figure}[t!]
	\centering
	\begin{subfigure}[b]{\textwidth}
		\centering
		\includegraphics[width=\linewidth]{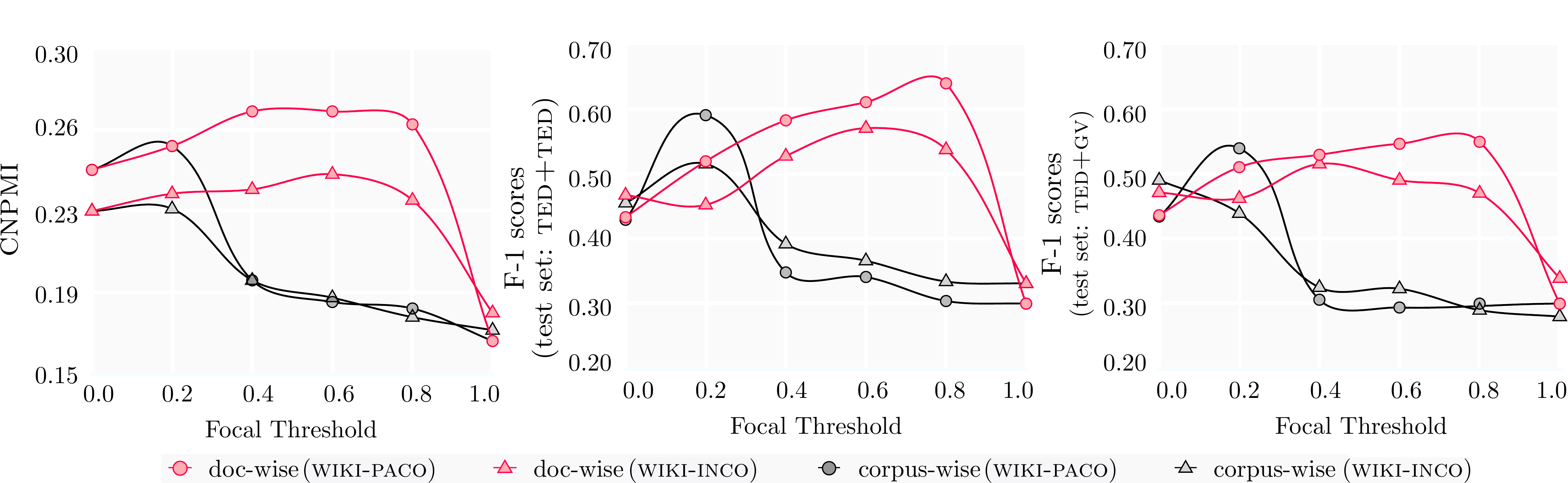}
		\caption{Training \softlink{} model.}
	\end{subfigure}
	\begin{subfigure}[b]{\textwidth}
		\centering
		\includegraphics[width=\linewidth]{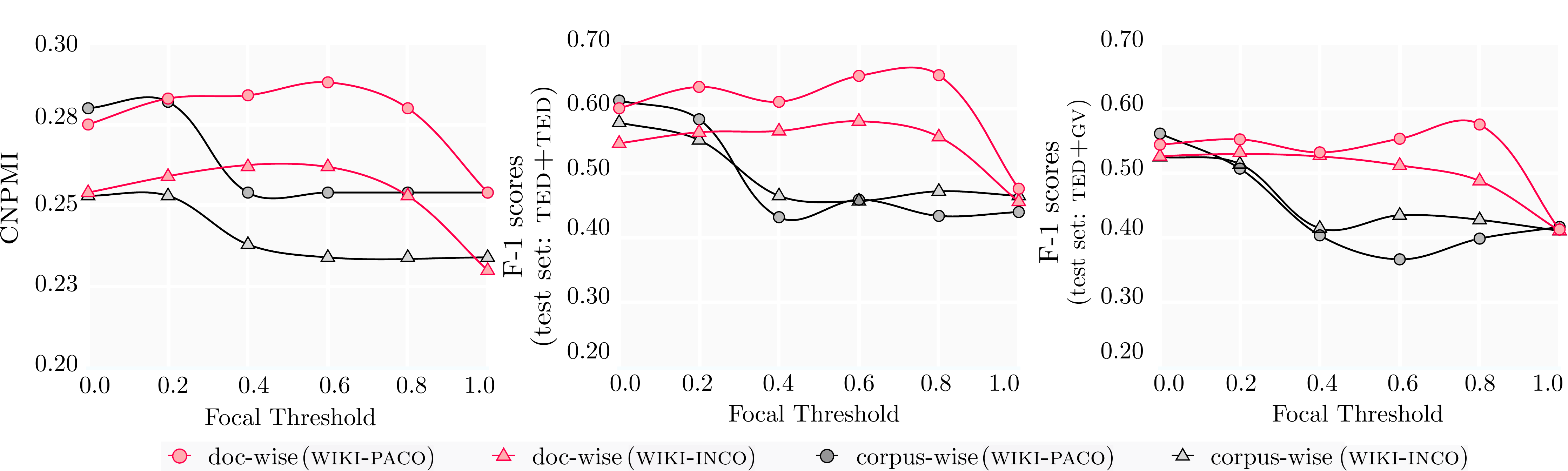}
		\caption{Training \softlink{} with \voclink{}.}
	\end{subfigure}
	\caption{\cnpmi{} scores and F-1 scores of crosslingual classification
		under different focal thresholds and selection scope of the transfer distribution for \softlink{} and \softlink{}+\voclink{} (Section~\ref{sec:threshold}).}
	\label{fig:clf}
\end{figure}

We have shown that, when optimized, \softlink{}
can better utilize dictionary resources and outperform
other models.
We now focus on different training configurations for \softlink{}, 
specifically, different methods of focusing the transfer distribution (Section~\ref{sec:threshold}).

Figure~\ref{fig:clf}
shows how F-1 and \cnpmi{} scores change with different static focusing methods.
We vary the focal threshold and selection scope (\textit{i.e.,} doc-wise or corpus-wise) for transfer distributions.
As we increase the focal threshold $\pi$,
more documents are zeroed out in the transfer distributions.
When $\pi=0.6$ or $0.8$,
the transfer distributions are very sparse,
and we notice that document-wise selection achieves the best performance.
In the extreme case that $\pi=1$,
the transfer distributions are all zero,
so \softlink{} loses its connections between $\ell_1$ and $\ell_2$,
and thus degrades to monolingual \lda{}.
When training \softlink{} with \voclink{},
the change of \cnpmi{} and F-1 scores are less obvious as we increase focal threshold,
since increasing focal threshold only has an impact on the \softlink{} component of the model.
When the focal threshold is higher, fewer soft links are active, so the model is closer to a plain \voclink{} model.

Interestingly, when focal threshold $\pi$ changes from $0.2$ to $0.4$,
F-1 scores of corpus-wise selection scope trained on \softlink{} drops
drastically, in contrast to document-wise.
This is because using corpus-wise selection
could set a large portion of transfer distributions to zero,
and only a small number of documents have non-zero transfer distributions.
Since corpus-wise selection relies on the entire training corpus,
it must be used with caution.

We find that using annealing to dynamically focus the distributions works well and is competitive with static focusing (Table~\ref{tbl:anneal}). 
Annealing does better than the majority of settings of static focusing, though is worse than optimally-tuned focusing.
We do not observe a significant difference between the two annealing schedules.
When combining \softlink{} and \voclink{},
the patterns are similar to that of \softlink{} only.

\begin{table}\centering\small
	\begin{subtable}{\textwidth}
		\centering
		\begin{tabular}{c|c|c|c|c|c|c}
		\hline 
		 & \multicolumn{2}{c|}{{F-1 scores (\ted{}+\ted{})}} & \multicolumn{2}{c|}{{F-1 scores (\ted{}+\gv{})}} & \multicolumn{2}{c}{{\cnpmi{}}}\\ 
		\cline{2-7} 
		& \lis{} & Fixed & \lis{} & Fixed & \lis{} & Fixed \\ \hline \hline
		{\wikipaco{}} & $0.627$ & $0.638$ & $0.551$ & $0.534$ & $0.256$ & $0.258$ \\ 
		\hline 
		{\wikiinco{}} & $0.551$ & $0.526$ & $0.475$ & $0.470$ & $0.220$ & $0.217$ \\ 
		\hline 
	\end{tabular}
	\caption{Training \softlink{} model.}
	\end{subtable}\\
	
\begin{subtable}{\textwidth}
	\centering
\begin{tabular}{c|c|c|c|c|c|c}
	\hline 
	 & \multicolumn{2}{c|}{{F-1 scores (\ted{}+\ted{})}} & \multicolumn{2}{c|}{{F-1 scores (\ted{}+\gv{})}} & \multicolumn{2}{c}{{\cnpmi{}}}\\ 
	\cline{2-7} 
	 & \lis{} & Fixed & \lis{} & Fixed & \lis{} & Fixed \\ \hline \hline
	{\wikipaco{}} & $0.640$ & $0.647$ & $0.557$ & $0.543$ & $0.261$ & $0.266$ \\ 
	\hline 
	{\wikiinco{}} & $0.546$ & $0.517$ & $0.459$ & $0.465$ & $0.242$ & $0.233$ \\ 
	\hline 
\end{tabular}
\caption{Training \softlink{} with \voclink{}.}
\end{subtable}
\caption{Dynamically focusing transfer distributions in \softlink{} yields competitive results on classification and topic quality evaluation. There is no significant difference between Fixed and \lis{} schedules.}
\label{tbl:anneal}
\end{table}

\subsection{Sensitivity to Dictionary Size}
\label{sec:dictsize}

Both \voclink{} and \softlink{} use the same dictionary resource,
yet \softlink{} produces better features for downstream tasks.
To understand this behavior better, we experiment with different dictionary sizes to understand how well the models are utilizing the resource.

\begin{figure}[t!]
	\centering
	\includegraphics[width=\linewidth]{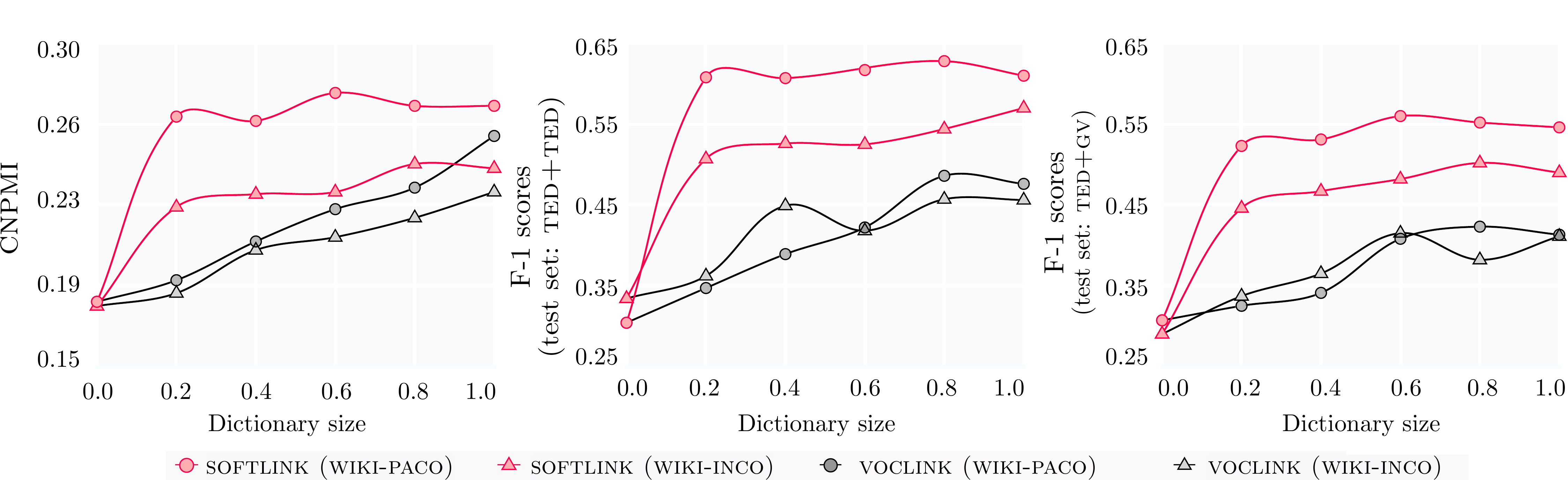}
	\caption{
		 Performance of \voclink{} continues increasing when more dictionary entries are added, while \softlink{} performance mostly stabilizes after using only $20\%$ of available dictionary entries.
	}
	\label{fig:size}
\end{figure}

In Figure~\ref{fig:size},
we use different proportions ($20\%$, $40\%$, $\ldots$, $80\%$) of the dictionary to train \softlink{} and \voclink{}.\footnote{We use document-wise selection scope and focal threshold $\pi=0.6$ for training \softlink{}; same as in Section~\ref{sec:baseline}.}
We observe that the performance of \voclink{} (both F-1 and \cnpmi{}) increases almost linearly with the dictionary size.
In contrast, \softlink{} is already at its best performance with only 20\% of the available dictionary entries. 
This is further confirmation that \softlink{} is using this resource in a more efficient way.

In \voclink{}, knowledge transfer happens through internal nodes of the word distribution priors, \textit{i.e.,}
word translations pairs,
and words without translations are directly connected to the Dirichlet tree's root.
If the dictionary cannot cover all the word types appeared in the training set,
\voclink{} will have a set of word types in $\ell_1$ that
cannot transfer enough topic knowledge to $\ell_2$ and vice versa.
The fewer entries the dictionary provides,
the more \voclink{} degrades to monolingual \lda{}.
In contrast, \softlink{}
can potentially transfer knowledge 
 from the whole corpus.
For \softlink{}, the dictionary is not used directly for modeling, rather it is only used for linking documents.
Thus, knowledge transfer does not heavily rely on the number
of entries in the dictionary.

\subsection{Discussion}

The sensitivity to dictionary size is an important factor
to be considered in practice.
For low-resource languages,
a dictionary is easier to obtain than a
 large parallel corpus (Section~\ref{sec:intro}).
Models that rely on dictionaries such as \voclink{}
and \softlink{} are therefore more applicable to low-resource languages than \hardlink{}.
However, there are also large variations in dictionary size among languages.
For example, in Wiktionary,
$57$ languages have fewer than $1{,}000$ entries,
while $77$ languages have more than $100{,}000$ entries.
For truly low-resource languages,
dictionary size could be a limiting factor.
Since \softlink{} can outperform \voclink{}
with only a limited amount of lexical information,
it may be able to transfer knowledge to low-resource languages more effectively than other approaches.

In summary,
\softlink{} relaxes and generalizes \hardlink{}
to be adaptable to more situations,
while using dictionary information more efficiently than \voclink{}.

\section{Conclusions and Future Work}

We have described a new formulation for \mltms{}
which explicitly shows the knowledge transfer process across languages.
Based on this analysis,
we proposed a new \mltm{}
that can learn multilingually coherent topics and
provide consistent topic features for crosslingual tasks.
Unlike existing models,
our approach is flexible and adaptable to incomparable corpora
with only a dictionary,
which is beneficial in many situations, in particular low-resource settings.

There are many possible directions following this work.
First,
our formulation of the knowledge transfer process
enables future work focusing on how to develop more efficient algorithms
that transfer knowledge with minimal supervision.
Second,
for \softlink{} 
we plan to explore more about characteristics of languages that can lead to better formulations and learning of the transfer distributions.

\bibliographystyle{acl}

\newpage
\setcounter{footnote}{0} 

\begin{appendices}
\section{Pseudolikelihood}

\setcounter{theorem}{0}
\begin{theorem}\label{th:pll}
	The conditional generative model with document links 
	yields the same posterior estimator to the joint generative model
	using collapsed Gibbs sampling.
\end{theorem}

\begin{proof}
Suppose the document links model is sampling topic of the $m$-th token in document $d_{\ell_2}$.
The sampler calculates the conditional topic distribution,
and then draw a topic assignment.
Using collapsed Gibbs sampling,
we calculate the conditional probability of a topic $k$:
\begin{equation*}
\resizebox{\textwidth}{!}{%
$\begin{aligned}
&\Pr\left(z_{\dt,m}=k|\vt{z}_{\dt,-},\vt{w}_\dt; \mathbf{n}_{d_{\ell 1}}, \alpha,  \beta\right)
= \frac{\Pr\left(z_{\dt,m}=k,\vt{z}_{\dt,-},\vt{w}_\dt; \mathbf{n}_{d_{\ell 1}}, \alpha,  \beta\right)}{\Pr\left(\vt{z}_{\dt,-},\vt{w}_\dt; \mathbf{n}_{d_{\ell 1}}, \alpha, \beta\right)} \\
=& \frac{\Pr\left(z_{\dt,m}=k,\vt{z}_{\dt,-},\vt{w}_\dt; \mathbf{n}_{d_{\ell 1}}, \alpha \right)}{\Pr\left(\vt{z}_{\dt,-}; \mathbf{n}_{d_{\ell 1}}, \alpha\right)} \cdot \frac{\Pr\left(\vt{w}_\dt|z_{\dt,n}=k,\vt{z}_{\dt,-};\beta\right)}{\Pr\left(\vt{w}_\dt|\vt{z}_{\dt,-},\beta\right)}\\
=&
\frac{\frac{\prod_{k'\neq k}\Gamma\left(n_{k'|\dt}+\mathbf{n}_{d_{\ell 1},k'}+\alpha\right)\cdot\Gamma\left(n_{k|\dt}+\mathbf{n}_{d_{\ell 1},k}+\alpha+1\right)}{\Gamma\left(n_{\cdot|d}+\mathbf{n}_{d_{\ell 1}}+\alpha+1\right)}}{\frac{\prod_k\Gamma\left(n_{k|\dt}+\mathbf{n}_{d_{\ell 1},k}+\alpha\right)}{\Gamma\left(n_{\cdot|\dt}+\mathbf{n}_{d_{\ell 1}}+K\alpha\right)}}
\cdot
\frac{\frac{\prod_{w\neq w_{\dt,m}}\Gamma\left(n_{w|k}+\beta\right)\cdot\Gamma\left(n_{w_{\dt,m}|k}+\beta+1\right)}{\Gamma\left(n_{\cdot|k}+V^{(\ell_2)}\beta+1\right)}}{\frac{\prod_w\Gamma\left(n_{w|k}+\beta\right)}{\Gamma\left(n_{\cdot|k}+V^{(\ell_2)}\beta\right)}}\\
=&
\frac{\Gamma\left(n_{k|\dt}+\mathbf{n}_{d_{\ell 1},k}+\alpha+1\right)}{\Gamma\left(n_{k|\dt}+\mathbf{n}_{d_{\ell 1},k}+\alpha\right)}
\cdot
\frac{\Gamma\left(n_{\cdot|\dt}+\mathbf{n}_{d_{\ell 1}}+K\alpha\right)}{\Gamma\left(n_{\cdot|\dt}+\mathbf{n}_{d_{\ell 1}}+\alpha+1\right)}
\cdot
\frac{\Gamma\left(n_{w_{\dt,m}|k}+\beta\right)}{\Gamma\left(n_{w_{\dt,m}|k}+\beta+1\right)}
\cdot
\frac{\Gamma\left(n_{\cdot|k}+V^{(\ell_2)}\beta\right)}{\Gamma\left(n_{\cdot|k}+V^{(\ell_2)}\beta+1\right)}
\\
=& \frac{n_{k|\dt}+\mathbf{n}_{d_{\ell 1},k}+\alpha}{n_{\cdot|\dt}+\mathbf{n}_{d_{\ell 1}}+K\alpha}
\cdot
\frac{n_{w_{\dt,m}|k}+\beta}{n_{\cdot|k}+V^{(\ell_2)}\beta},\label{gibbs}
\end{aligned}$%
}
\end{equation*}
where 
$\vt{z}_{\dt,-}$ is all the topic assignments in $d_{\ell_2}$ except the current one,
$n_{\cdot|d_{\ell 2}}$ the number of tokens in $d_{\ell_2}$,
$n_{k|d_{\ell 2}}$ the number of tokens assigned to topic $k$ in $d_{\ell_2}$,
$n_{\cdot|k}$ the number of tokens assigned to topic $k$,
$n_{w|k}$ the number of word type $w$ assigned to topic $k$, and
$V^{(\ell_2)}$ the vocabulary size of language $\ell_2$.
The roles of $\ell_1$ and $\ell_2$ are interchangeable,
so both languages use the same conditional distributions.
The last equation of the derivation above gives identical posterior estimation
in the original model.
Thus, the alternative formulation,
despite not a numerically accurate likelihood approximation,
does not make a difference for parameter estimation.
\end{proof}

\section{Dataset Processing Details}

\subsection{Pre-Processing}

For all the languages,
we use existing stemmers to stem words in the corpora and the entries in Wiktionary.
Since Chinese does not have stemmers,
we loosely use ``stem'' to refer to ``segment'' Chinese sentences into words.
We also use fixed stopword lists to filter out stop words.
Table \ref{tools} lists the source of the stemmers and stopwords.

\begin{table*}
	\centering
	\begin{tabular}{c|c|c|c}
		\hline 
		Language & Family & Stemmer & Stopwords \\ \hline \hline
		\en{} & Germanic & \texttt{SnowBallStemmer}~\footnotemark &  NLTK \\ \hline 
		\es{} & Romance & \texttt{SnowBallStemmer} & NLTK  \\ \hline 
		\ru{} & Slavic & \texttt{SnowBallStemmer} &  NLTK \\ \hline 
		\ar{} & Semitic & \texttt{Assem's Arabic Light Stemmer}~\footnotemark &  GitHub~\footnotemark \\ \hline 
		\fa{} & Indo-Iranian & \texttt{Hazm}~\footnotemark &  GitHub \\ \hline
		\zh{} & Sinitic & \texttt{Jieba}~\footnotemark &  GitHub \\ \hline 
	\end{tabular}
	\caption{List of source of stemmers and stopwords used in experiments.}
	\label{tools}
\end{table*}

\footnotetext[1]{\url{http://snowball.tartarus.org};}
\footnotetext[2]{\url{http://arabicstemmer.com};}
\footnotetext[3]{\url{https://github.com/6/stopwords-json};}
\footnotetext[4]{\url{https://github.com/sobhe/hazm};}
\footnotetext[5]{\url{https://github.com/fxsjy/jieba}.}

\subsection{Data Source}

We list the statistics in Table~\ref{tbl:stats}.

\begin{table}
\small
	\centering
\begin{tabular}{r|r|r|r|r|r|r|r}\hline
	& & \wikipaco & \wikiinco & \ted & \gv & \begin{tabular}[x]{@{}c@{}}Wikipedia\\(for \cnpmi{})\end{tabular} & Wiktionary \\ \hline\hline
	\multirow{3}{*}{ \ar } & \#docs & 2,000 & 2,000 & 1,112 & 2,000 & 8,862 & \multirow{3}{*}{ 16,127 } \\
	& \#tokens & 1,075,691 & 293,640 & 1,521,334 & 466,859 & 79,740 & \\
	& \#types & 32,843 & 19,900 & 44,982 & 32,468 & 1,533,261 & \\ \hline
	\multirow{3}{*}{ \es } & \#docs & 2,000 & 2,000 & 1,152 & 2,000 & 9,325 & \multirow{3}{*}{ 31,563 } \\
	& \#tokens & 475,234 & 237,561 & 1,228,469 & 493,327 & 1,763,897 & \\
	& \#types & 35,069 & 27,465 & 30,247 & 28,471 & 91,428 & \\ \hline
	\multirow{3}{*}{ \fa } & \#docs & 2,000 & 2,000 & 687 & 401 & 9,669 & \multirow{3}{*}{ 14,952 } \\
	& \#tokens & 415,620 & 91,623 & 1,415,263 & 89,414 & 940,672 & \\
	& \#types & 18,316 & 9,987 & 36,670 & 9,447 & 46,995 & \\ \hline
	\multirow{3}{*}{ \ru } & \#docs & 2,000 & 2,000 & 1,010 & 2,000 & 9,837 & \multirow{3}{*}{ 33,574 } \\
	& \#tokens & 4,368,563 & 766,887 & 1,133,098 & 679,217 & 2,356,994 & \\
	& \#types & 51,740 & 24,341 & 44,577 & 47,395 & 134,424 & \\ \hline
	\multirow{3}{*}{ \zh } & \#docs & 2,000 & 2,000 & 1,123 & 2,000 & 8,222 & \multirow{3}{*}{ 23,276 } \\
	& \#tokens & 3,095,977 & 303,634 & 1,428,532 & 745,307 & 1,338,116 & \\
	& \#types & 59,431 & 30,481 & 71,906 & 69,872 & 144,765 & \\ \hline
\end{tabular} 
	\caption{Statistics of corpora and dictionary in the five languages used in the experiments.}
	\label{tbl:stats}
\end{table}

\paragraph{Wikipedia (\wikipaco{}, and \wikiinco{}).}

For training multilingual topic models,
the dataset Wikipedia can be downloaded at \url{http://opus.nlpl.eu/TED2013.php}.
For each language pair (\en{}, $\ell$),
we create \wikiinco{}, a completely incomparable corpus,
where $2,000$ \en{} documents and $2,000$ non-English documents
are randomly chosen but do not contain document-level translations to each other.

We also create \wikipaco{},
a partially comparable corpus.
Each language has different proportions of comparable document pairs.
See Table~\ref{tbl:prop}.
\begin{table}\centering
\begin{tabular}{c|ccccc}
	\hline
	Languages & \ar{} & \es{} & \fa{} & \ru{} & \zh{}  \\ \hline\hline
	Proportion & 12.2\% & 9.35 \% & 50.85 \% & 50.20 \% & 17.90 \% \\
	\hline
\end{tabular}
\caption{Proportions of linked document pairs in corpus \wikipaco{}.}
\label{tbl:prop}
\end{table}

\paragraph{TED Talks 2013 (\ted{}).}

TED Talks 2013 contains mostly parallel documents,
and can be obtained from OPUS:
\url{http://opus.nlpl.eu/TED2013.php}.
Note that not all English documents have translations to another language,
which is slightly different from the original assumptions in polylingual topic models.

The classification labels can be obtained from the documents.
Each document has several ``categories'' that can be regarded as labels.
Thus, we retrieve those labels, and choose the most frequent five labels for classification:
\textit{technology}, \textit{culture}, \textit{science}, \textit{global issues}, and \textit{design}.

\paragraph{Global Voices (\gv{}).}

Global Voices can be obtained from OPUS as well: \url{http://opus.nlpl.eu/GlobalVoices.php}.
Global Voices corpus has a large number of documents,
so for efficiency, we randomly choose a sample of at most $2{,}000$ documents for each language.

There's no label information from the corpus itself.
However, the labels can be retrieved from the webpage of each document, at \url{https://globalvoices.org}.
To make sure Global Voices have the same label set to TED Talks,
we changed the label set to: 
\textit{technology}, \textit{culture}, \textit{science}, \textit{business}, and \textit{politics}.

\paragraph{Wiktionary.}

We use English Wiktionary to create bilingual dictionaries,
which can be downloaded at \url{https://dumps.wikimedia.org/enwiktionary/}.

\end{appendices}

\end{document}